\documentclass[conference]{IEEEtran}
\IEEEoverridecommandlockouts

\usepackage{url}
\usepackage{color}
\usepackage{bm}
\usepackage{booktabs}
\usepackage{algorithm}
\usepackage{algorithmicx}
\usepackage{paralist}
\usepackage{mathtools}
\usepackage[noend]{algpseudocode}
\usepackage{colortbl}
\usepackage{verbatim}
\definecolor{mygray}{gray}{.9}

\usepackage{cite}
\usepackage{amsmath,amssymb,amsfonts}
\usepackage{graphicx}
\usepackage{textcomp}
\usepackage{xcolor}
\def\BibTeX{{\rm B\kern-.05em{\sc i\kern-.025em b}\kern-.08em
    T\kern-.1667em\lower.7ex\hbox{E}\kern-.125emX}}
\begin{document}

\title{Hierarchical Bayesian Personalized Recommendation: A Case Study and Beyond}

\author{\IEEEauthorblockN{Zitao Liu}
\IEEEauthorblockA{\textit{TAL AI Lab} \\
\textit{TAL Education Group}\\
Beijing, China \\
liuzitao@100tal.com}
\and
\IEEEauthorblockN{Zhexuan Xu}
\IEEEauthorblockA{\textit{Intelligent Advertising Lab} \\
\textit{JD.COM}\\
Mountain View, CA USA \\
zhexuan.xu1@jd.com}
\and
\IEEEauthorblockN{Yan Yan}
\IEEEauthorblockA{\textit{Facebook Inc} \\
Menlo Park, CA USA \\
chrisyan@fb.com}
}

\maketitle

\begin{abstract}
Items in modern recommender systems are often organized in hierarchical structures. These hierarchical structures and the data within them provide valuable information for building personalized recommendation systems. In this paper, we propose a general hierarchical Bayesian learning framework, i.e., \emph{HBayes}, to learn both the structures and associated latent factors. Furthermore, we develop a variational inference algorithm that is able to learn model parameters with fast empirical convergence rate. The proposed HBayes is evaluated on two real-world datasets from different domains. The results demonstrate the benefits of our approach on  item recommendation tasks, and show that it can outperform the state-of-the-art models in terms of precision, recall, and normalized discounted cumulative gain. To encourage the reproducible results, we make our code public on a git repo: \url{https://tinyurl.com/ycruhk4t}.

\end{abstract}


\section{Introduction}
\label{sec:intro}
Real-world organizations in business domains operate in a multi-item and multi-level environment. Items and their corresponding information collected by these organizations often reflect a hierarchical structure. For examples, products in retail stores are usually stored in hierarchical inventories. News on web pages is created and placed hierarchically in most websites. These hierarchical structures and the data within them provide a large amount of information when building effective recommendation systems. Especially in the e-commerce domain, all products are displayed in a site-wide hierarchical catalog and how to build an accurate recommendation engine on top of it becomes one of the keys to majority companies' business success. 

However, how to utilize the rich information behind hierarchical structures to make personalized and accurate product recommendations still remains challenging due to the unique characteristics of hierarchical structures and the modeling trade-offs arising from them. Briefly, most well-established recommendation algorithms cannot naturally take hierarchical structures as additional inputs. Moreover, flattening hierarchical structures usually doesn't work well. It will not only blow up the entire feature space but introduce noise when training the recommendation models. On the other hand, discarding hierarchies will lead to recommendation inaccuracies. The most common way to alleviate this dilemma is to feed every piece of data from the hierarchy into a complex deep neural network and hope the neural network itself can figure out a way to intelligently utilize the hierarchical knowledge. However, such approaches usually behave more like black boxes which brings much difficulty to debug and cannot provide any interpretation of the intermediate results or outcomes.

In this work, we propose and develop a hierarchical Bayesian modeling framework, a.k.a., \emph{HBayes}, that is able to flexibly capture various relations between items in hierarchical structures from different recommendation scenarios. By introducing latent variables, all hierarchical structures are encoded as conditionally independences in HBayes graphical models. Moreover, we develop a variational inference algorithm for efficiently learning parameters of HBayes. 

To illustrate the power of the proposed HBayes approach, we introduce HBayes by first using a real-world apparel garment recommendation problem as an example. As an illustration, we generalize apparel styles, product brands and apparel items into a three-level hierarchy, and add additional latent variables as the apparel style membership variables to capture the diverse and hidden style properties of each brand.  Furthermore, we include user-specific features into HBayes and extend the model into the supervised learning settings where user feedback events such as clicks and conversions are incorporated.  Note that the HBayes framework is not only limited to apparel recommendation. In the end, we show its flexibility and effectiveness on another music recommendation problem as well.

Overall this paper makes contributions in four folds:

\begin{itemize}
\item It presents a generalized hierarchical Bayesian learning framework to learn from rich data with hierarchies in real cases.
\item It provides a variational inference algorithm that can learn the model parameters with very few iterations.
\item It evaluates HBayes and its benefits comprehensively in tasks of apparel recommendation on a real-world data set. 
\item It tests the HBayes framework in different recommendation scenarios to demonstrate the model generalization.
\end{itemize}

The remainder of the paper is organized as follows: Section \ref{sec:related} provides a review of existing recommendation algorithms and their extensions in hierarchal learning settings.  Section \ref{sec:method} introduces the notations and our generalized HBayes learning framework and its variational inference algorithm. In Section \ref{sec:experiment}, we conduct experiments in a real-world e-commerce data set to show the effectiveness of our proposed recommendation algorithm in different aspects. In addition, we test our model on a music recommendation data set to illustrate the generalization and extended ability of HBayes. We summarize our work and outline potential future extensions in Section \ref{sec:conclusion}.

\section{Related Work}
\label{sec:related}
Collaborative filtering \cite{Sarwar:2001:ICF:371920.372071,Su:2009:SCF:1592474.1722966}, matrix factorization \cite{Rendle:2010:FPM} and content-based recommendation approaches \cite{2011rsh..book...73L,Liu:2011,Yuan:2015} are well studied and these approaches have become the foundations of recommender systems. A group of studies have been done to improve the existing performance. Tang, Hu, and Liu take time as an additional input dimension, aiming at explicitly modeling user interests over time \cite{Tang_review:2013}.  Koren et al. develops a collaborative filtering type of approach with predictions from static average values combining with a dynamic changing factor \cite{Koren:2010}. Yin et al. proposes a user-tag-specific temporal interest model to track user interests over time by maximizing the time weighted data likelihood \cite{Yin:2011}.  

Recently, there are works using Bayesian inferencing for recommendation tasks.  Rendle et al. combines the Bayesian inference and matrix factorization together for learning users implicit feedbacks (click \& purchase) that is able to directly optimize the recommendation ranking results \cite{rendle2009bpr}. Ben-Elazar et al. and Zhang et al.  take user preference consistency into account and develop a variational Bayesian personalized ranking model for better music recommendation \cite{Ben-Elazar:2017,zhang2007efficient}.  However, these approaches do not leverage the item structural information when building their Bayesian models. Given the fact that the hierarchical structural information widely exists in real-world recommendation scenarios such as e-commerce, social network, music, etc. failing to utilize such information makes these Bayesian approaches inefficient and inaccurate.  

Hierarchical information is a powerful tree-based structure that encodes human knowledge. Several research works have been conducted to utilize this information to improve recommendations. For examples, Shepitsen et al. rely on the hierarchies generated by user-taggings to build a better personalized recommender system \cite{shepitsen2008personalized}.  Wang et al. introduce a hierarchical matrix factorization approach that exploits the intrinsic structural information to alleviate cold-start and data sparsity problems \cite{wang2018exploring}.  Despite the fact that these hierarchical recommender systems have received some success, there are still challenges such as: (1) how to infer the hierarchical structure efficiently and accurately if it is not explicit? (2) how to better understand the hierarchical topologies discovered by recommendation approaches? and (3) how to utilize the inferred hierarchical information for precise data explanations?




\section{The HBayes Framework}
\label{sec:method}
In this work, we develop our generalized hierarchical Bayesian modeling framework that is able to capture the hierarchical structural relations and latent relations in the real-world recommendation scenarios.  Note that we take apparel recommendation as a case study for the ease of model description, but our model framework is general enough to be enforced to other hierarchical data recommendation cases.

\subsection{Generative Process}

In the real-world scenario, each item or product has to come with a brand and a brand may have more than one items in the hierarchical structures. Therefore, we denote each event $t$ as a 4-tuple (\emph{Item}, \emph{Brand}, \emph{User}, \emph{IsClick}), i.e, ($\mathbf{X}_t, b_t, u_t, y_t$). $\mathbf{X}_t$ represents the item features associated with event $t$ and $y_t$ is the binary label that indicates whether user $u_t$ has clicked $\mathbf{X}_t$ or not. $b_t$ is the brand of item $\mathbf{X}_t$. 

Furthermore, we expand the hierarchy by a hidden factor, i.e., ``style''. Products from each brand $b_t$ tend to exhibit different styles or tastes, which are unknown but exist. In this paper, brands are represented as random mixtures over latent styles, where each style is characterized by a distribution over all the items. Let $S$, $B$, $U$ and $N$ be the total number of styles, brands, users and events.

The generative process of HBayes can be described as follows:

\begin{enumerate}
\item[\textbf{Step 1.}] Draw a multivariate Gaussian prior for each user $k$, i.e, $\mathbf{U}_k \sim \mathcal{N}(\mathbf{0}, \delta_u^{-1}\mathbf{I})$ where $k \in \{1, \cdots, U\}$.

\item[\textbf{Step 2.}] Draw a multivariate Gaussian prior for each style $j$, i.e, $\mathbf{S}_j \sim \mathcal{N}(\mathbf{w}, \delta_s^{-1}\mathbf{I})$ where $j \in \{1, \cdots, S\}$.

\item[\textbf{Step 3.}] Draw a style proportion distribution $\boldsymbol{\theta}$ for each brand $i$, $\boldsymbol{\theta} \sim \mbox{Dir}(\boldsymbol{\gamma})$  where $i \in \{1, \cdots, B\}$.

\item[\textbf{Step 4.}] For each brand $i$:
	\begin{enumerate}
		\item[\textbf{Step 4.1}] Draw style assignment $\mathbf{z}_i$ for brand $i$ where the selected style $p$ is sampled from $\mbox{Mult}(\boldsymbol{\theta})$. $\mathbf{z}_i$ is a $S\times1$ one hot encoding vector that $z_{i,p} = 1$ and $z_{i,j} = 0$ for $j = 1, \cdots, p-1, p+1, \cdots, S$.
		\item[\textbf{Step 4.2}] Draw $\mathbf{B}_i \sim \mathcal{N}(\mathbf{S}_{p}, \delta_b^{-1}\mathbf{I})$.
	\end{enumerate}

\item[\textbf{Step 5.}] For each event $t$, draw $y_t$ from Bernoulli distribution where the probability $p$ is defined as $p(y_t|\mathbf{x}_t, \mathbf{B}_{b_t}, \mathbf{U}_{u_t})$.
\end{enumerate}

\noindent where $\delta_s$, $\delta_u$ and $\delta_b$ are the scalar precision parameters and $\mathbf{w}$ is the prior mean of $\mathbf{S}_j$. $Dir(\cdot)$ and  $Mult(\cdot)$ represent Dirichlet distribution and multinomial distribution, respectively. 

With consideration of model flexibility and capacity, we also treat each distribution's parameter as a random variable and define hyper-priors on top. More specifically, We draw the prior mean $\mathbf{w}$ from $\mathcal{N}(\mathbf{0}, \delta_w^{-1}\mathbf{I})$. For $\delta_w$, $\delta_s$, $\delta_u$ and $\delta_b$, we define Gamma priors over them i.e.,  $p(\delta_*) = \mathcal{G}(\alpha, \beta)$, where $\delta_* \in \{ \delta_w, \delta_s, \delta_u, \delta_b \}$. 

\begin{figure}[htb]
\centering
\includegraphics[width=0.75\linewidth]{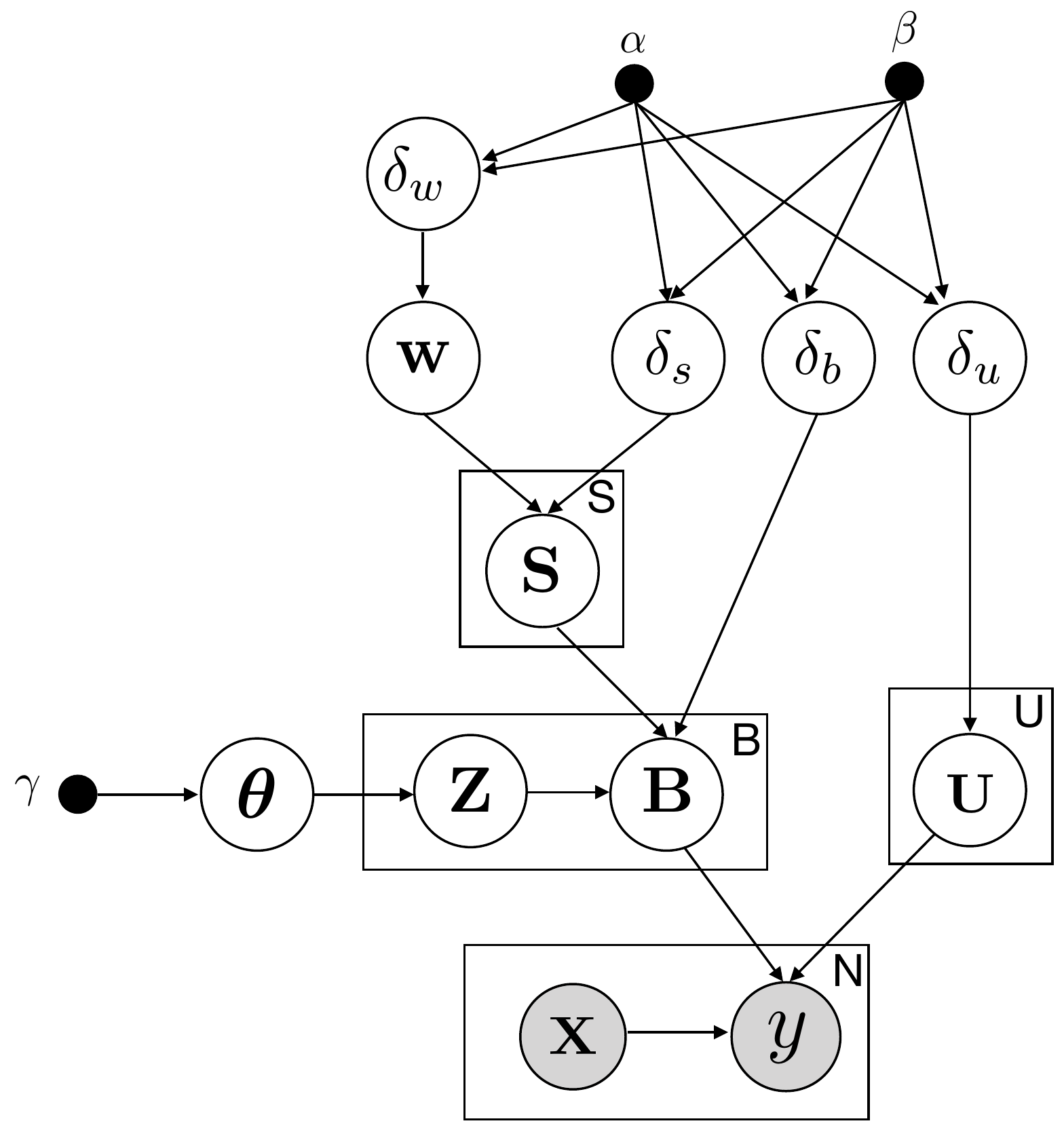}
\caption{A graphical model representation of HBayes.}
\label{fig:model}
\end{figure}

The family of probability distributions corresponding to this generative process is depicted as a graphical model in Figure \ref{fig:model}.

\subsection{Probability Priors \& Models}

In this work, we model the probability of a single event $t$ given ($\mathbf{X}_t, b_t, u_t, y_t$) as 

\begin{equation}
\label{eq:sigmoid_prob}
p\big(y_t|\bm{X}_t,\bm{B}_{b_t},\bm{U}_{u_t} \big)= \sigma^{y_t}\big(h_t\big) \cdot \big(1-\sigma\big(h_t\big)\big)^{1-y_t}
\end{equation}

\noindent where $\sigma(\cdot)$ is a logistic function, i.e. $\sigma(x)=(1+e^{-x})^{-1}$. $h_{t}=\bm{X}_t^T(\bm{B}_{b_t}+\bm{U}_{u_t})$. $\mathbf{X}^T$ is the vector transpose of $\mathbf{X}$. $\bm{U}_{u_t}$ represents user specific information encoded in HBayes for user $u_t$ and $\bm{B}_{b_t}$ denotes the brand $b_t$'s specific information.

As mentioned in Step 3 of the generative process of HBayes, each brand $i$'s style proportion distribution $\boldsymbol{\theta}$ follows a Dirichlet distribution: $p(\boldsymbol{\theta}) \sim Dir(\boldsymbol{\gamma})$, where $\bm{\gamma}$ is the $S$-dimensional Dirichlet hyper-parameter. We initialize $\gamma_j$ by $\frac{1}{S}$. 

Furthermore, in Step 4 of the generative process of HBayes, a brand is modeled as a random mixture over latent styles. Hence, we model the brand parameters by a mixture of multivariate Gaussian distribution defined as follows:

\begin{equation}
p(\bm{B}_i|\bm{z}_i,\bm{S},\delta_b) = \prod_{j}^S \mathcal{N}(\bm{B}_i; \bm{S}_j,\delta_b^{-1}\mathbf{I}) ^{\mathbb{I}(z_{i,j}=1)}
\end{equation}

\noindent where $z_{i,j}$ is the \emph{j}th element of $\mathbf{z}_i$ and $\mathbb{I}(\xi)$ is an indicator function that $\mathbb{I}(\xi)=1$ if the statement $\xi$ is true; $\mathbb{I}(\xi)=0$ otherwise.

Therefore, the log joint likelihood of the dataset $\mathcal{D}$, latent variable $\bm{Z}$ and the parameter $\Theta$ by given hyper-parameters $\mathcal{H} = \{\gamma, \alpha,\beta\}$ could be written as follows:

\begin{align}
 & \log \big( p(\mathcal{D},\bm{Z},\Theta|\mathcal{H}) \big) \nonumber \\
= & \sum_{t=1}^N \log p(y_t|\bm{X}_t,\bm{B}_{b_t},\bm{U}_{u_t}) + \sum_{i=1}^B \log p(\bm{B}_i|\bm{z}_i,\bm{S},\delta_b)  \nonumber \\
+ & \sum_{i=1}^B \log p(\bm{z}_{i}|\bm{\theta}) + \sum_{j=1}^S \log p(\bm{S}_j|\bm{w},\delta_s) + \log p(\bm{\theta}|\bm{\gamma})  \nonumber  \\
+ & \sum_{k}^U \log p(\bm{U}_k|\delta_u) + \log p(\bm{w}|\delta_w) + \log p(\delta_w|\alpha,\beta) \nonumber \\
+ & \log p(\delta_u|\alpha,\beta) + \log p(\delta_b|\alpha,\beta) + \log p(\delta_s|\alpha,\beta)
\label{eq:log_likelihood}
\end{align}

We use $\Theta$ to denote all model parameters, i.e., $\Theta = \Big\{\{\bm{U}_k\}, \{\bm{B}_i\}, \{\bm{S}_j\}, \bm{w}, \boldsymbol{\theta}, \delta_u,\delta_b,\delta_s,\delta_w \Big\}$, where $k \in \{1, \cdots, U\}$, $i \in \{1, \cdots, B\}$, $j \in \{1, \cdots, S\}$.

\subsection{Optimization}

Since both $\mathbf{Z}$ and $\Theta$ defined by HBayes are unobserved, we cannot learn HBayes directly. Instead, we infer the expectations of these latent variables and compute the expected log likelihood of the log joint probability with respect to the latent variables distribution, i.e., $\mathcal{Q}$ function defined in eq.(\ref{eq:original_q}). In the following, we omit the explicit conditioning on $\mathcal{H}$ for notational brevity. 

\begin{equation}
\label{eq:original_q}
\mathcal{Q} = \int_\Theta \sum_{\bm{Z}} p(\bm{Z}, \Theta| \mathcal{D})  \log \big( p(\mathcal{D},\bm{Z},\Theta) \big) d\Theta
\end{equation}

From the Bayes rule, the posteriors distribution of $\mathbf{Z}$ and $\Theta$ can be represented by $p(\bm{Z},\Theta|\mathcal{D}) = \frac{p(\mathcal{D},\bm{Z},\Theta)}{p(\mathcal{D})}$. However, this above distribution is intractable to compute in general \cite{dickey1983multiple}. To tackle this problem, a wide variety of approximation inference algorithms are developed, such as Laplace approximation \cite{rue2009approximate}, variational approximation \cite{bishop2006pattern}, and Markov Chain Monte Carlo (MCMC) approach \cite{blei2003latent}, etc.

In this work, we choose to solve this problem by using variational Bayes approximation \cite{bishop2006pattern}. More specifically, we approximate the original posterior distribution $p(\bm{Z}, \Theta| \mathcal{D})$ with a tractable distribution $q(\bm{Z}, \Theta)$ such that instead of maximizing the $\mathcal{Q}$ function defined in eq.(\ref{eq:original_q}), we maximize the variational free energy defined as 

\begin{equation}
\label{eq:original_variational_energy}
\mathcal{Q}'(q) = \int_\Theta \sum_{\bm{Z}} q(\bm{Z},\Theta) \log\frac{p(\mathcal{D},\bm{Z},\Theta)}{q(\bm{Z},\Theta)}d\Theta
\end{equation}

\noindent which is also equal to minimize the KL divergence of $p(\bm{Z}, \Theta| \mathcal{D})$ and $q(\bm{Z}, \Theta)$.

Here we choose to apply \emph{Mean Field} approximation technique to approximate $p(\bm{Z}, \Theta| \mathcal{D})$, where we assume independence among all different variables ($\mathbf{Z}$ and $\Theta$) and define $q(\bm{Z}, \Theta)$ as follows:

\begin{align}
\label{eq:mean_field}
q(\bm{Z}, \Theta) = & q(\mathbf{Z}) \cdot \prod_{k=1}^K q(\mathbf{U}_k) \cdot \prod_{j=1}^S q(\mathbf{S}_j) \cdot \prod_{i=1}^B q(\mathbf{B}_i)  \nonumber \\
& \cdot q(\mathbf{w}) \cdot q(\boldsymbol{\theta}) \cdot q(\delta_u) \cdot q(\delta_b) \cdot q(\delta_s) \cdot q(\delta_w)
\end{align}

\noindent where $q$ denotes different distribution functions for notation brevity. Details of choices of different distributions will be discussed in Section \ref{sec:param}.

\subsubsection{Sigmoid Approximation}

The Gaussian priors from our log joint probability (see eq.(\ref{eq:log_likelihood})) are not conjugate to the data likelihood due to the fact that our events are modeled by a sigmoid function (see eq.(\ref{eq:sigmoid_prob})). In order to conduct tractable inference on $\mathcal{Q}'(q)$, we apply a variational lower bound approximation on eq.(\ref{eq:sigmoid_prob}) that has the ``squared exponential'' form. Therefore, they are conjugate to the Gaussian priors.

\begin{equation*}
\sigma(h_t) \geq \sigma(\xi_t)\exp\big\{\frac{1}{2}(h_t-\xi_t)-\lambda_t(h_t^2-\xi_t^2)\big\}
\end{equation*}

\noindent where $\lambda_t=\frac{1}{2\xi_t}[\sigma(\xi_t)-\frac{1}{2}]$ and $\xi_t$ is a variational parameter. This lower bound is derived using the convex inequality. The similar problem was discussed in \cite{jaakkola1997variational,jordan1999introduction}.

Therefore, each event likelihood can be expressed as follows:

\begin{align}
\label{eq:likelihood_approx}
& \sigma^{y_t}(h_t) \cdot  \big(1-\sigma(h_t) \big)^{1-y_t} = \exp\big( y_t h_t \big) \sigma(-h_t) \nonumber \\
\geq & \sigma(\xi_t)\exp \big(y_t h_t-\frac{1}{2}(h_t+\xi_t)-\lambda_t(h_t^2-\xi_t^2) \big)
\end{align}

By using the sigmoid approximation in eq.(\ref{eq:likelihood_approx}), our variational free energy $\mathcal{Q}'(q)$ (eq.(\ref{eq:original_variational_energy})) can be bounded as:

\begin{equation}
\label{eq:approx_variational_energy}
\mathcal{Q}'(q) \geq \mathcal{Q}'_{\xi}(q) = \int_\Theta \sum_{\bm{Z}} q(\bm{Z},\Theta) \log\frac{p_{\xi}(\mathcal{D},\bm{Z},\Theta)}{q(\bm{Z},\Theta)}d\Theta
\end{equation}

In the following, we will maximize the lower bound of the variational free energy $\mathcal{Q}_{\xi}'(q) $ for parameter estimation.

\subsubsection{Parameter Estimation}
\label{sec:param}

We develop a Variational Bayes (VB) algorithm for HBayes parameter estimation, where in the E-step, we compute the expectation of the hidden variables $\mathbf{Z}$ and in the M-step, we try to find $\Theta$ that maximizes lower bound of the variational free energy $\mathcal{Q}_{\xi}'(q)$ (eq.(\ref{eq:approx_variational_energy})). In the VB algorithm, we use coordinate ascent variational inference (CAVI) \cite{bishop2006pattern} to optimize $\mathcal{Q}_{\xi}'(q)$. CAVI iteratively optimizes each factor of the mean field variational distribution, while holding the others fixed.

\noindent \textbf{Update expectation of $Z$}:
We assume each brand's style membership latent variable is independent and therefore, $q(\mathbf{Z}) = \prod_{i=1}^B q(\mathbf{z}_i)$. For each $\mathbf{z}_i$, we parameterize $q(\mathbf{z}_i)$ and update $\mu_{i,j}$ based on the multinomial distribution:

$$q(\mathbf{z}_i) = \prod_{j=1}^S \mu_{i,j}^{\mathbb{I}(z_{i,j}=1)}; ~~~~\mu_{i,j} =  \frac{\rho_{i,j}}{\sum_{p=1}^S\rho_{i,p}}$$


\begin{align}
\ln(\rho_{i,j}) & =  \mathbb{E}[\ln(\bm{\theta}_j)]+\frac{d}{2}\mathbb{E}[\ln(\delta_b)]-\frac{d}{2}\ln(2\pi) \nonumber \\ 
- & \frac{1}{2}\mathbb{E}\big[\delta_b(\mathbf{B}_i-\mathbf{S}_j)^\mathrm{T}(\mathbf{B}_i-\mathbf{S}_j)\big]
\end{align}

\noindent where the expectation $\mathbb{E}[\cdot]$ is with respect to the (currenctly fixed) variational density over $\Theta$  i.e., $\mathbb{E}[\cdot]=\mathbb{E}_{-\bm{Z}}[\cdot]$ in this part. Furthermore, in the following, we note that: 
\begin{equation}
\label{eq:hidden}
\mathbb{E}[z_{i,j}]=\mu_{i,j}
\end{equation}

\noindent \textbf{Parametrization and update rule of $q(\mathbf{\theta})$}:
For the style proportion distribution $\mathbf{\theta}$, we parameterize $q(\mathbf{\theta})$ as a Dirichlet distribution, i.e., $q(\mathbf{\theta}) = Dir(\mathbf{\theta}; \mathbf{\gamma})$, and the update rule for $\gamma$ are 

\begin{equation}
\label{eq:start}
\gamma_j = \gamma_j + \sum_{i=1}^B \mu_{i,j}, j=1,\cdots,S
\end{equation}

\noindent \textbf{Parametrization and update rule of $q(\mathbf{U}_k)$}:
For each user $k$, $k = 1, \cdots, U$, we parameterize $q(\mathbf{U}_k)$ as a multivariate normal distribution, i.e., $q(\mathbf{U}_k) = \mathcal{N}(\mathbf{U}_k; \bm{\mu}^u_k, \bm{\Sigma}_k^u)$, and the update rule for $\bm{\mu}^u_k$, $\bm{\Sigma}_k^u$ are 

\begin{align}
\bm{\Sigma}_k^u = \big[  \delta_u \mathbf{I} + \sum_{t=1}^N \mathbb{I}(u_t = k) 2\lambda_t \mathbf{X}_t \mathbf{X}_t^T  \big]^{-1} ~~~~~~~~~~~\\ 
\hskip -0.5cm \bm{\mu}^u_k = \bm{\Sigma}_k^u\Big[  \sum_{t=1}^N\mathbb{I}(u_t=k)  \big(y_t-\frac{1}{2}-2\lambda_{t}\mathbf{X}_t^T\mathbb{E}[\mathbf{B}_{b_t}]\big)\mathbf{X}_t \Big]
\end{align}

\noindent \textbf{Parametrization and update rule of $q(\mathbf{B}_i)$}:
For each brand $i$, $i = 1, \cdots, B$, we parameterize $q(\mathbf{B}_i)$ as a multivariate normal distribution, i.e., $q(\mathbf{B}_i) = \mathcal{N}(\mathbf{B}_i; \bm{\mu}^b_i, \bm{\Sigma}^b_i)$, and the update rule for $\bm{\mu}^b_i$, $\bm{\Sigma}^b_i$ are 

\begin{align}
\bm{\Sigma}^b_i & = \Big[\delta_b\sum_{j=1}^S\mu_{i,j}\mathbf{I}+\sum_{t=1}^N\mathbb{I}(b_t=i)2\lambda_{t}\mathbf{X}_t\mathbf{X}_t^T\Big]^{-1} \\ 
\bm{\mu}^b_i & = \bm{\Sigma}_i^b\Big[  \delta_b\sum_{j=1}^S\mu_{i,j}\mathbb{E}[\mathbf{S}_j] \nonumber \\ 
+ & \sum_{t=1}^N\mathbb{I}(b_t=i)\big(y_t-\frac{1}{2}- 2\lambda_{t}\mathbf{X}_t^T\mathbb{E}[\mathbf{U}_{u_t}]\big)\bm{X}_t \Big]
\end{align}

\noindent \textbf{Parametrization and update rule of $q(\mathbf{S}_j)$}:
For each style $j$, $j = 1, \cdots, S$, we parameterize $q(\mathbf{S}_j)$ as a multivariate normal distribution, i.e., $q(\mathbf{S}_j) = \mathcal{N}(\mathbf{S}_j; \bm{\mu}^s_j, \bm{\Sigma}^s_j)$, and the update rule for $\bm{\mu}^s_j$, $\bm{\Sigma}^s_j$ are 

\begin{align}
\bm{\Sigma}^s_j = \big[\delta_s+\delta_b\sum_{i=1}^{B}\mu_{i,j}\big]^{-1}\mathbf{I};~~~ \bm{\mu}^s_j = \bm{\Sigma}_j^{s}\Big[ \delta_s\mathbb{E}[\bm{w}]+\delta_b\sum_{i=1}^B\mu_{i,j}\mathbb{E}[\mathbf{B}_i] \Big]
\end{align}

\noindent \textbf{Parametrization and update rule of $q(\mathbf{w})$}:
For the mean variable of style prior $\mathbf{w}$, we parameterize $q(\mathbf{w})$ as a multivariate normal distribution, i.e., $q(\mathbf{w}) = \mathcal{N}(\mathbf{w}; \bm{\mu}^w, \bm{\Sigma}^w)$, and the update rule for $\bm{\mu}^w$, $\bm{\Sigma}^w$ are 

\begin{align}
\bm{\Sigma}^w  = \big[\delta_w+\delta_s\cdot S\big]^{-1}\mathbf{I};~~~\bm{\mu}^w = \bm{\Sigma}^w\Big[ \delta_s\sum_{j=1}^S\mathbb{E}[\mathbf{S}_j] \Big]
\end{align}

\noindent \textbf{Parametrization and update rule of $q(\delta_u)$, $q(\delta_b)$, $q(\delta_s)$ and $q(\delta_w)$}:
For all the precision parameters' distributions, we parameterize them as a Gamma distribution, i.e., $p(\delta_*) = \mathcal{G}(\delta_*; \alpha_*, \beta_* )$, where $\delta_* \in \{ \delta_w, \delta_s, \delta_u, \delta_b \}$ and the update rule for are $\alpha_\mathrm{new} = \alpha_\mathrm{old} + \Delta \alpha$ and $\beta_\mathrm{new} = \beta_\mathrm{old} + \Delta \beta$, separately:

\begin{align}
\Delta\alpha_u &= \frac{dU}{2},~  \Delta\beta_u=\frac{1}{2}\sum_{k=1}^{U}\mathbb{E}\big[\mathbf{U}_k^T\mathbf{U}_k\big] \nonumber\\ 
\Delta\alpha_b &= \frac{dB}{2},  \Delta\beta_b=\frac{1}{2}\sum_{i=1,j=1}^{B,S}\mu_{i,j}\mathbb{E}\big[(\mathbf{B}_i-\mathbf{S}_j)^T(\mathbf{B}_i-\mathbf{S}_j)\big] \nonumber\\
\Delta\alpha_s &= \frac{dS}{2},  \Delta\beta_s=\frac{1}{2}\sum_{j=1}^{S}\mathbb{E}\big[(\mathbf{S}_j-\mathbf{w})^T(\mathbf{S}_j-\mathbf{w})\big] \nonumber\\
\Delta\alpha_w &= \frac{d}{2},~  \Delta\beta_w=\frac{1}{2}\mathbb{E}\big[\mathbf{w}^T\mathbf{w}\big] 
\end{align}

\noindent \textbf{Update rule of $\xi$}:
For the variational parameters $\xi_{t}, t=1,\cdots,N$, in order to maximize $\mathcal{Q}'_{\xi}(q)$ such that the bound on $\mathcal{Q}'(q)$ is tight \cite{bishop2006pattern}, the update rule is: 

\begin{equation}
\label{eq:end}
\xi_{t}=\sqrt{\mathbb{E}\Big[\big(\bm{X}_t^T(\mathbf{B}_{b_t}+\mathbf{U}_{u_t}\big)^2\Big]}
\end{equation}

\subsubsection{Summary}
The parameter estimation method for the HBayes is summarized by Algorithm \ref{alg:hbayes}.

\begin{algorithm}
\caption{Parameter Estimation in HBayes}
\label{alg:hbayes}
\begin{algorithmic}[1]
\State{\small \textbf{INPUT}:}
\State{Hyper-parameters $\mathcal{H}$:  $\mathcal{H} = \{\alpha, \beta, \boldsymbol{\gamma}\}$}
\State{Data samples $\mathcal{D}$: $(\mathbf{X}_t, b_t, u_t, y_t)$, $t = 1, \cdots N$}

\Procedure{Learning HBayes}{}
\Repeat
\State E-step: compute expectation of $\mathbf{Z}$ by eq.(\ref{eq:hidden}).
\State M-step: estimate $\{\bm{U}_k\}$, $\{\bm{B}_i\}$, $\{\bm{S}_j\}$, $\bm{w}$, $\boldsymbol{\theta}$, $\delta_u$,$\delta_b,\delta_s,\delta_w,\xi_{t}$ by eq.(\ref{eq:start}) - eq.(\ref{eq:end}).
\Until{Convergence}
\State
\Return $\Theta$
\EndProcedure
\end{algorithmic}
\end{algorithm}

\subsection{Prediction}

In the recommender system, the task is to generate the top $K$ product list for each user.  Given the user $u^*$,  it's straightforward to expose top M products based on the probability of the positive outcomes. For the $m^{th}$ item, the probability is calculated as:

\begin{equation}
\label{prediction}
\begin{split}
\hat{y}_m & = p(y_m=1|\bm{X}_m,\mathcal{D},\mathcal{H}) \approx \int \sigma(h_{m})q(\bm{Z},\Theta)\mathrm{d}\Theta \nonumber \\
& = \int \sigma(h_{m})\mathcal{N}(h_{m}|\mu_{m},\sigma^2_{m})\mathrm{d}h_{m} \approx \sigma(\frac{\mu_{m}}{\sqrt{1+\pi\sigma^2_{m}/8}})
\end{split}
\end{equation}

\noindent where $h_{m}$ is a random variable with Gaussian distribution:

\begin{align*}
 h_{m} & =\bm{X}_m^T(\bm{B}_{b_m}+\bm{U}_{u^*})  \sim \mathcal{N}(h_{m};\mu_{m},\sigma^2_{m}) \\ 
\mu_{m} & =\mathbb{E}\big[\bm{X}_m^T(\bm{B}_{b_m}+\bm{U}_{u^*})\big] \\
\sigma^2_{m} & =\mathbb{E}\Big[\big(\bm{X}_m^T(\bm{B}_{b_m}+\bm{U}_{u^*})-\mu_{m}\big)^2\Big]
\end{align*}

\section{Experiments}
\label{sec:experiment}
In this section, we conduct several experiments on two data sets: (1) our case study: the real-world e-commerce apparel data set; (2) the publicly available music data set.  For both data sets, we compare HBayes against several other \emph{state-of-the-art} recommendation approaches which are briefly mentioned as follows:

\begin{itemize}
\item \textbf{HSR} \cite{wang2015exploring} explores the implicit hierarchical structure of users and items so the user preference towards certain products is better understood.  

\item \textbf{HPF} \cite{gopalan2015scalable} generates a hierarchical \emph{Poisson} factorization model to capture each user's latent preference.  Unlike proposed HBayes, HPF does not leverage the entity content feature for constructing the hierarchical structure.  

\item \textbf{SVD++} \cite{mnih2008probabilistic,koren2008factorization} combines the collaborative filtering and latent factor approaches.  

\item \textbf{CoClustering} \cite{george2005scalable} is a collaborative filtering approach based on weighted co-clustering improvements.  

\item \textbf{Factorization Machine (FM)} \cite{rendle2010factorization,rendle2012factorization} combines support vector machines with factorization models. Here, we adopt the LibFM implementation mentioned in \cite{rendle2012factorization} specifically as another baseline. 

\item \textbf{LambdaMART} \cite{burges2010ranknet} is the boosted tree version of LambdaRank \cite{donmez2009local}, which is based on RankNet \cite{burges2005learning}.  
\end{itemize}

\subsection{Evaluation Metrics}
Throughout the experiments, we compare HBayes against other baselines on the testing held-out dataset under the 5-fold cross-validation settings.  For each fold, after fitting the model on the training set, we rank on the testing set by each model, and generate the top K samples with maximal ranking scores for recommendation.  Regarding metrics, we adopt the \textbf{precision} and \textbf{recall} for evaluating the retrieval quality and normalized discounted information gain (NDCG) for evaluating the recommendation ranking quality which is defined as:  


\begin{equation}
\text{DCG@K} = \sum_{i=1}^K\frac{r_i}{\log_2(i+1)}; \text{NDCG@K} = \frac{\text{DCG@K}}{\text{IDCG@K}} \nonumber \\
\end{equation}

\subsection{Recommendation on Apparel Data}
The first apparel data set is collected from a large e-commerce company.  In this dataset, each sample represents a particular apparel product which is recorded by various features including: categories, titles, and other properties, etc.  Meanwhile, the user click information is also recorded and translated into data labels. Throughout the experiment, positive labels indicate that certain recommended products are clicked by the user, whereas negative samples indicate that the recommended products are skipped by the user which usually implies that the user is `lack of interest' towards that certain item.  By data cleaning and preprocessing: (1) merging duplicated histories; (2) removing users of too few records, the post-processed data set ends up with 895 users, 81223 products, 5535 brands with 380595 uniquely observed user-item pairs.  In average, each user has 425 products records, ranging from 105 to 2048, and 61.2\% of the users have fewer than 425 product clicking records.  For each item, we encode the popularity and category features into a 20 dimensional feature vector; title and product property into a 50 dimensional feature vector.  Combining with all features, the total dimension of each sample ends up with 140.

\subsubsection{Feature Analysis}
The apparel data are composed of four types of features: (1) product popularity; (2) product category; (3) product title; (4) product properties.  We briefly explain each feature's physical meaning and how we process the data as follows:

\begin{enumerate}

\item \textbf{Product Popularity} (POP): product popularity is a measure of the prevalence of certain items in the dataset. In general, customers have preference for a particular product during a period of time. This phenomenon is pretty common for apparel products (\emph{e.g.} apparels' popularity in certain styles may be affected by certain people or events, especially by those important public figures). For a particular product $i$, the popularity is defined as: $\mathrm{POP}_{i} \coloneqq \frac{n_{x_i}}{\mathcal{N}_\mathbf{x}}$, where $n_{x_i}$ are the number of orders or the contribution of gross merchandise volume (GMV) for product $i$,  and $\mathcal{N}_\mathbf{x} = \sum_{\forall x_i}n_{x_i}$ is the summation of $n_{x_i}$ across all products in the dataset. 

\item \textbf{Product Category} (CID): In e-commerce, items are clustered into different groups based on the alliance/ similarity of the item functionalities and utilities.  In e-commerce websites, such an explicit hierarchy usually exists.  We encode each item's category into a high dimensional vector via one-hot encoding and adjust the feature weights by the popularity of such category. 

\item \textbf{Product Title} (TITLE): product titles are created by vendors and they are typically in the forms of natural languages indicating the item functionality and utility.  Examples could be like `\emph{INMAN short sleeve round neck triple color block stripe T-shirt 2017}'.  We preprocess the product titles by generating the sentence embedding based on \cite{de2016representation}.  The main idea is to average the wording weights in the title sentence based on the inverse document frequency (IDF) value of each individual word involved. 

\item \textbf{Product Property Features} (PROP): other product meta-data features are also provided in the apparel dataset. For instance, the color feature of items takes values such like: `black', `white', `red', etc, and the sizing feature takes values such like: `S', `M', `L', `XL', etc.  Similar to category features, product properties are first encoded into binary vectors $\bm{x}_i \in \{0, 1\}^{|N|}$, where $N$ denotes the set of all possible product values.  Then the property binary vectors are hashed into fixed length ($\mathbf{50}$) vectors. 

\end{enumerate}

On one hand, by utilizing more features HBayes in general reaches better performance in terms of precision-recall metrics.  We report PR-AUC in Table \ref{tab:features_cmp} to prove this argument; on the other hand, we need to balance the features dimension and experiment performance considering that more features need more training time.  The experiment showed that the model spends less than 10 minutes to converge by only taking POP feature while needs more than 4 hours for POP+CID+TITLE+PROP features. \footnote{The experiment is conducted via the same Linux Qual core 2.8 GHz Intel Core i7 MacBook with 16 Gigabytes of memory.}.

\begin{table}[htb]
\centering
\caption{Model performance under different feature combinations in terms of PR AUC.}
\label{tab:features_cmp}
\begin{tabular}{l|c}
\toprule
\textbf{Features} & \textbf{PR AUC} \\
\hline
POP & 0.0406 \\
\rowcolor{mygray}
POP+CID & 0.0414 \\
POP+CID+TITLE & 0.0489 \\
\rowcolor{mygray}
POP+CID+TITLE+PROP & 0.0491 \\
\bottomrule
\end{tabular}
\end{table}

\subsubsection{Performance Comparison}
\begin{figure*}[!hbpt]
\center
\includegraphics[width=0.75\linewidth]{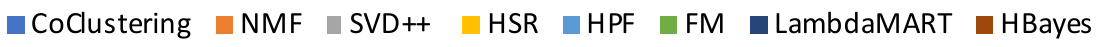}
\minipage{0.5\textwidth}
  \includegraphics[width=\linewidth]{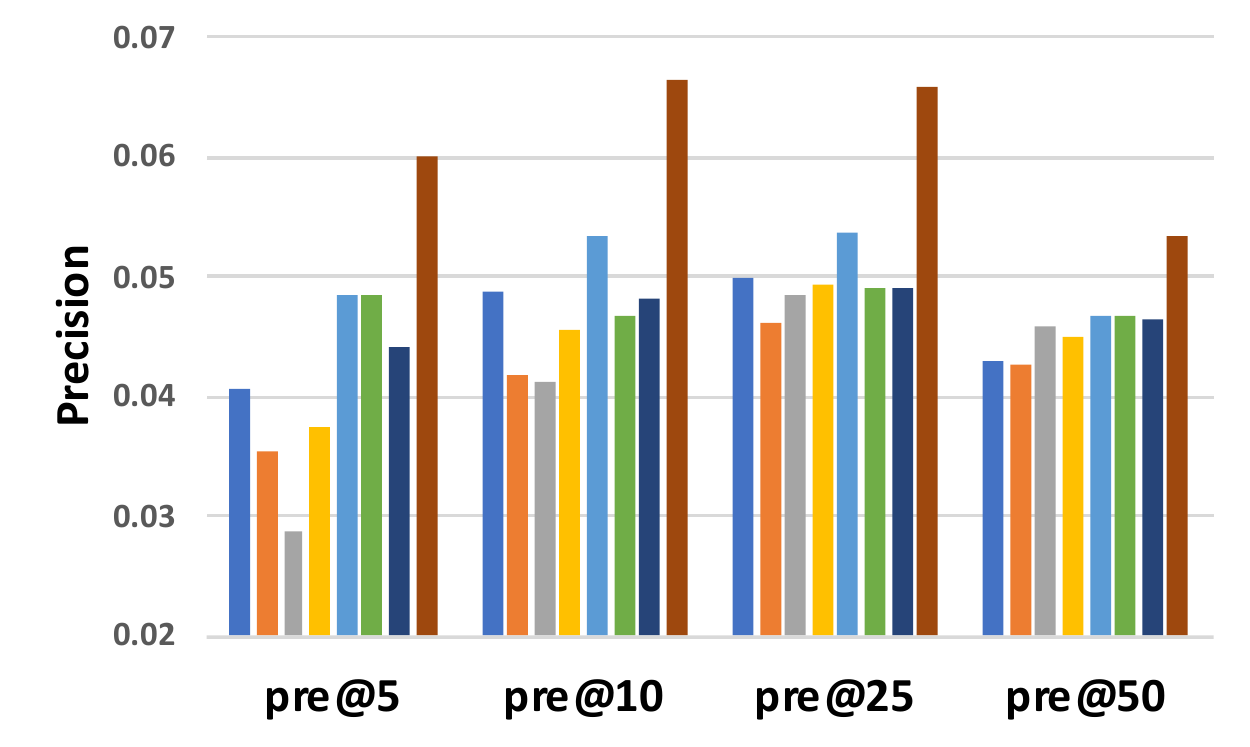}
  \caption{Precision@K on \emph{Apparel}.}
  \label{fig:perf_cmp_precision_apparel}
\endminipage\hfill
\minipage{0.5\textwidth}%
  \includegraphics[width=\linewidth]{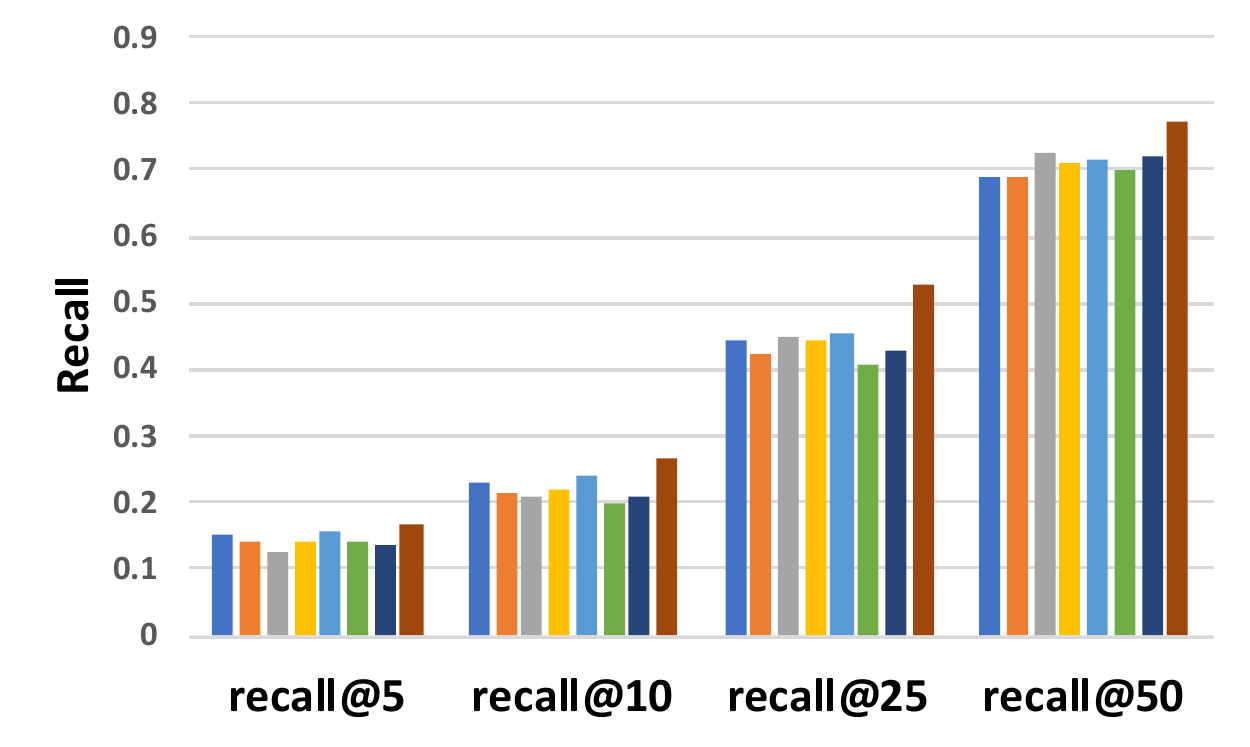}
  \caption{Recall@K on \emph{Apparel}.}
  \label{fig:perf_cmp_recall_apparel}
\endminipage
\end{figure*}

We first report the model performance regarding precision and recall for HBayes against other baselines in Figure \ref{fig:perf_cmp_precision_apparel} and Figure \ref{fig:perf_cmp_recall_apparel}.  As shown, when each method recommends fewer number of products (\emph{K} = 5), HBayes does not show the superiority regarding with recalls, with the increment of recommended items, the recall for HBayes becomes much better against others, which implies HBayes is really efficient in terms of finding items that people tend to take interest in.  In the sense of precision, HBayes is consistently better than other baseline methods which implies HBayes is much more accurate in terms of item classification under different \emph{K}. 

Regarding the ranking quality, we use NDCG to report each method's performance in Table \ref{tab:NDCG_cmp_apparel}.  HBayes is superior against other baseline methods through out different K recommended.  Specially, HBayes beats the second best HPF at $\text{K}=5$ by $10.3\%$, at $\text{K}=10$ by $12.4\%$, at $\text{K}=25$ by $14.7\%$ and at $\text{K}=50$ by $11.2\%$.  

\begin{table}[!phtb]
\begin{center}
\caption{NDCG on apparel recommendations.}
\label{tab:NDCG_cmp_apparel}
\begin{tabular}{l|cccc}
\toprule
\textbf{Method} & \textbf{NDCG@5} & \textbf{NDCG@10} & \textbf{NDCG@25} & \textbf{NDCG@50} \\
\hline
\rowcolor{mygray}
CoClustering & 0.1288 & 0.1637 & 0.2365 & 0.3050 \\
NMF & 0.1249 & 0.0156 & 0.2272 & 0.3020 \\
\rowcolor{mygray}
SVD++ & 0.1138 & 0.1487 & 0.2287 & 0.3073\\
HSR & 0.1266 & 0.1603 & 0.2354 & 0.3107 \\
\rowcolor{mygray}
HPF & 0.1412 & 0.1757 & 0.2503 & 0.3229 \\
FM & 0.1363 & 0.1592 & 0.2291 & 0.3117\\
\rowcolor{mygray}
LambdaMART & 0.1287 & 0.1585 & 0.2304 & 0.3123\\
HBayes & \textbf{0.1557} & \textbf{0.1974} & \textbf{0.2871} & \textbf{0.3590}\\
\bottomrule
\end{tabular}
\end{center}
\end{table}

\subsubsection{Model Learning Analysis}
\begin{figure}[!htb]
\includegraphics[width=\linewidth]{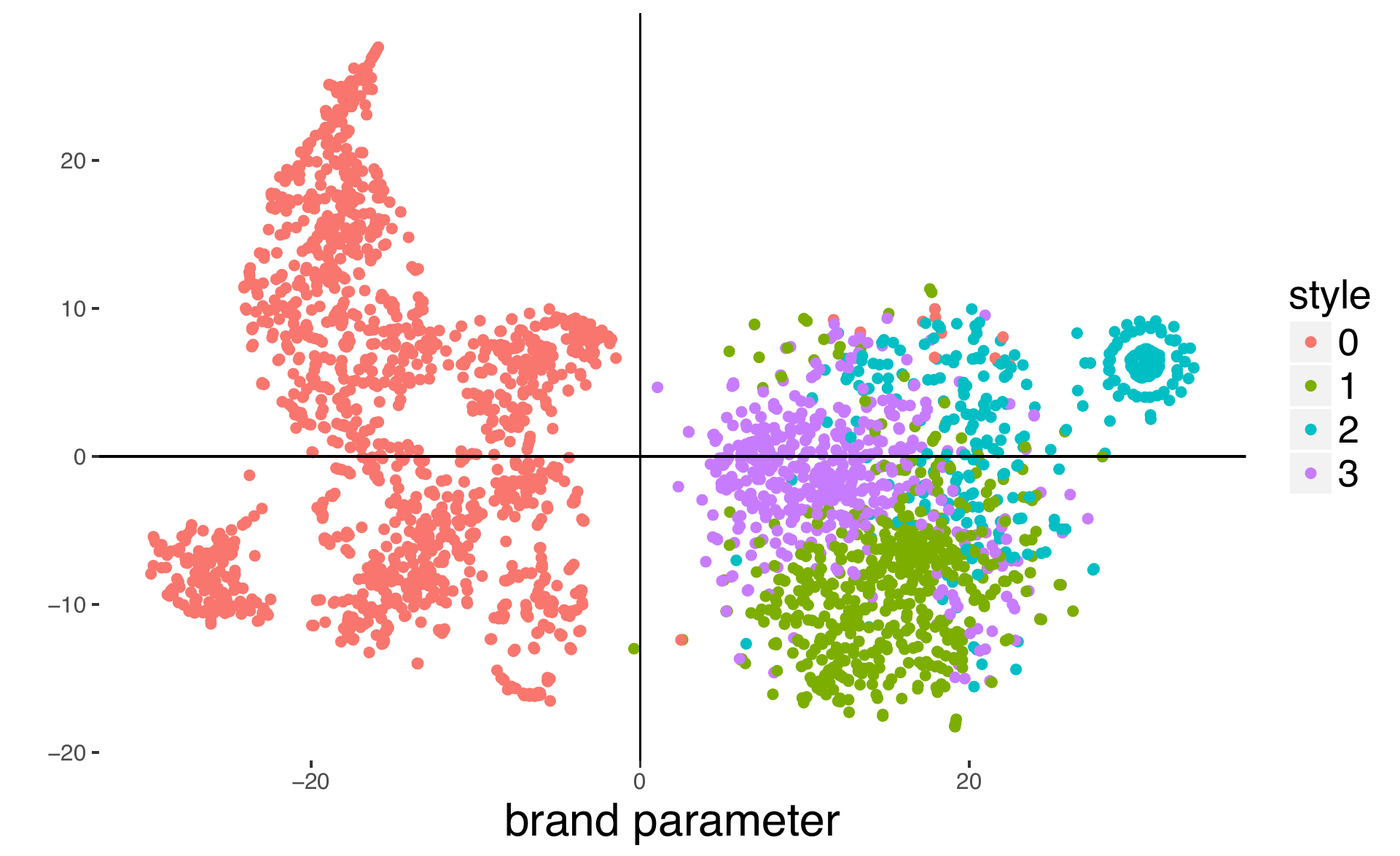}
\caption{tSNE of latent style clusters.}
\label{fig:tsne-represntation-style-cluster}
\end{figure}

\begin{figure}[!htb]
\includegraphics[width=0.85\linewidth]{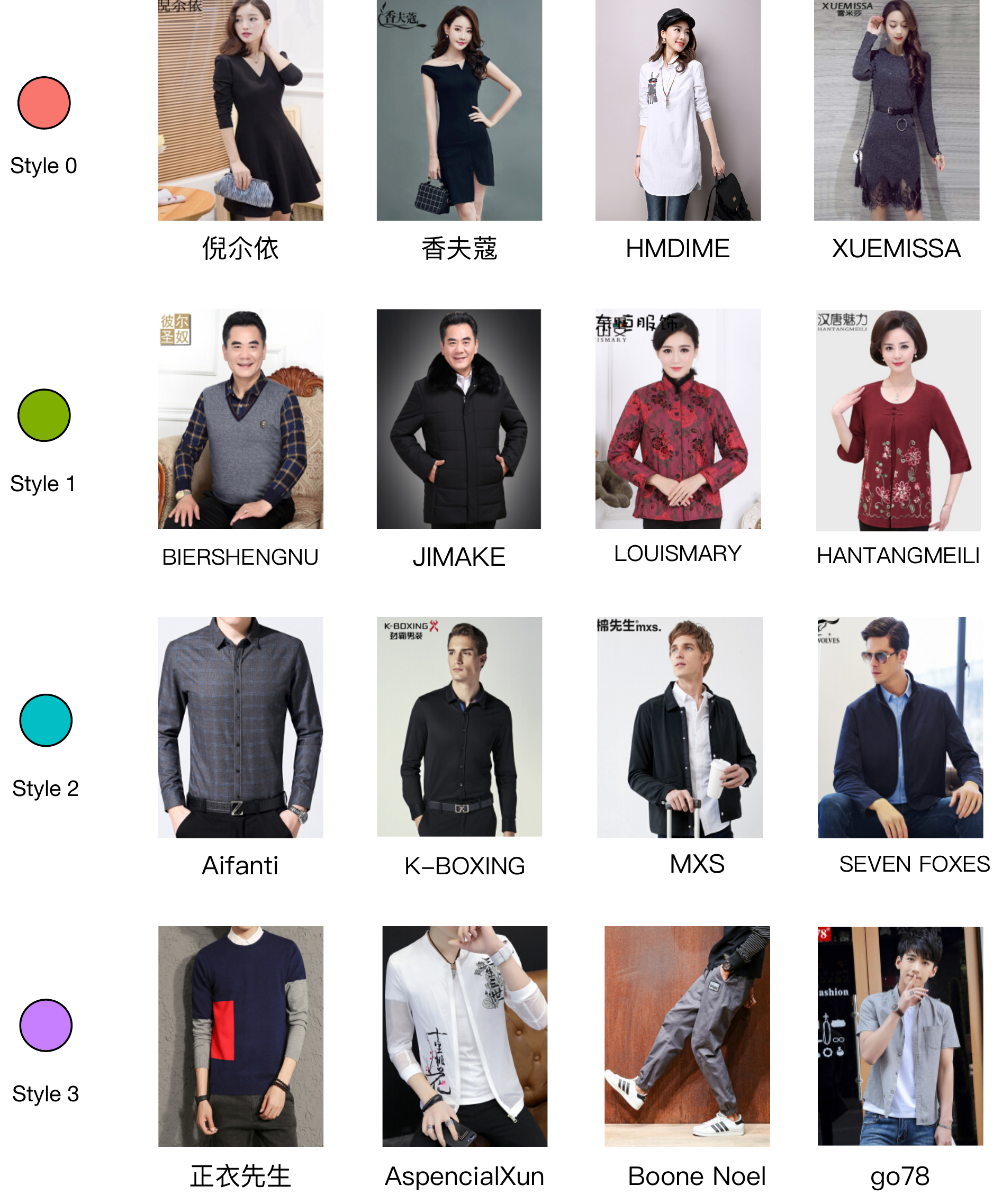}
\caption{Apparel examples in each style clusters.}
\label{fig:style-cluster-example}
\end{figure}

HBayes learns the latent style clusters, and group different brands of products based on their different hidden style representations. Figure \ref{fig:tsne-represntation-style-cluster} shows the tSNE \cite{maaten2008visualizing} representations of different apparel clusters learned by HBayes. We randomly pick $4$ samples out of each cluster and display each product image in Figure \ref{fig:style-cluster-example}.  As shown, cluster one which takes the majority proportion of apparel items seems about stylish female youth garment.  This intuitively makes sense because the majority of apparel customers are young females for e-commerce websites; as a result, most apparels are focusing on the young female audience as well. The second cluster seems about senior customers who are elder in age. Interestingly, the third cluster and the fourth cluster that are closely tied up are both about young male customers.  However, the third cluster seems focusing more on office business garment while the fourth cluster seems more about Korean-pop street styles.  This indicates us that the HBayes indeed learns the meaningful intrinsic garment styles from apparel items by leveraging customer behavior data.

\subsection{Recommendation on Last.fm Music Data}

The second data set is collected from Last.fm dataset \cite{Celma:Springer2010} and Free Music Archive (FMA) \cite{FMA}. Last.fm is a publicly available dataset which contains the whole listening habits (till May, 5th 2009) for 1000 users. FMA is an open and easily accessible dataset providing 917 GiB and 343 days of Creative Commons-licensed audio from 106574 tracks, 16341 artists and 14854 albums, arranged in a hierarchical taxonomy of 161 genres. It also provides full-length and high-quality audios with precomputed features.  In our experiment, tracks in Last.fm dataset were further intersected with FMA dataset for better feature generation. The resulting dataset contains 500 users, 16328 tracks and 36 genres.

\subsubsection{Performance Comparison}
\begin{figure*}[!hbpt]
\center
\includegraphics[width=0.75\linewidth]{legend}
\minipage{0.5\textwidth}
  \includegraphics[width=\linewidth]{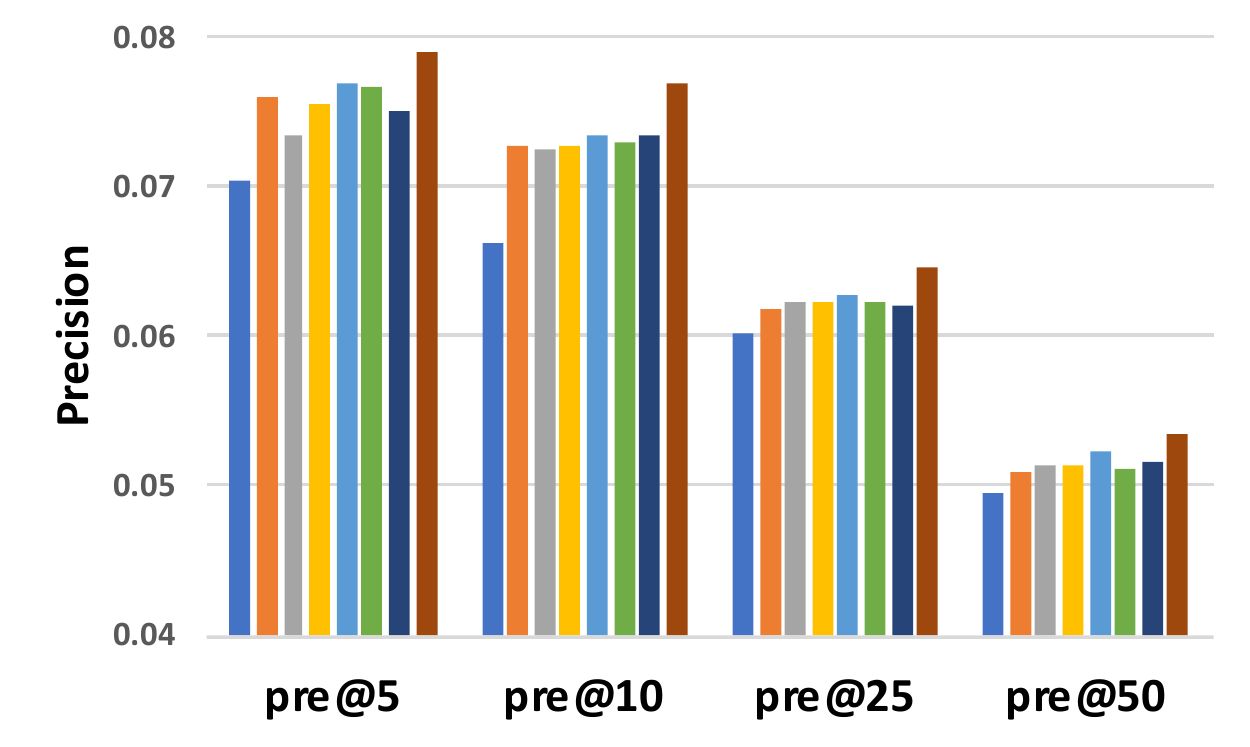}
  \caption{Precision@K on \emph{Music}.}
  \label{fig:perf_cmp_precision_music}
\endminipage\hfill
\minipage{0.5\textwidth}%
  \includegraphics[width=\linewidth]{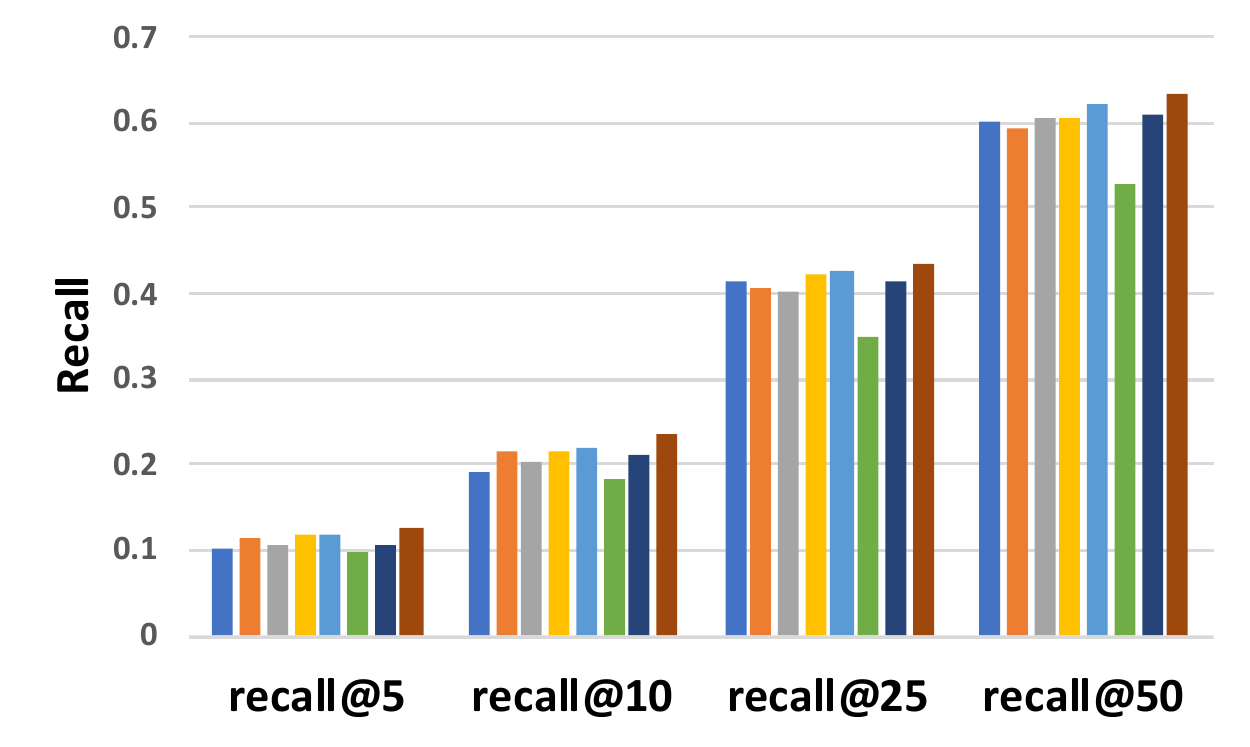}
  \caption{Recall@K on \emph{Music}.}
  \label{fig:perf_cmp_recall_music}
\endminipage
\end{figure*}

We conduct similar experiments as we do for apparel dataset and report precision and recall in Figure \ref{fig:perf_cmp_precision_music} and Figure \ref{fig:perf_cmp_recall_music}. Although HPF and HBayes share similar performance regarding recalls along with different \emph{K}, HBayes is dominant for precisions at different \emph{K}, especially when \emph{K} is small (5, 10), which indicates HBayes is very efficient and precise for helping users pick up the songs they prefer even when the recommended item lists are short. 

For ranking qualities, we report the NDCG performance in Table \ref{NDCG_cmp_music}.  Similar as e-commerce apparel data, HBayes is the best approach and HPF is the second best one throughout different K items recommended.  Specifically, HBayes beats HPF for $3.9\%$ at $\text{K}=5$, $4.1\%$ at $\text{K}=10$, $3.0\%$ at $\text{K}=25$, and $2.7\%$ at $\text{K}=50$ separately.

\begin{table}[htb]
\begin{center}
\caption{NDCG results on Last.fm recommendations.}
\label{NDCG_cmp_music}
\begin{tabular}{l|cccc}
\toprule
\textbf{Method} & \textbf{NDCG@5} & \textbf{NDCG@10} & \textbf{NDCG@25} & \textbf{NDCG@50} \\
\hline
\rowcolor{mygray}
CoClustering & 0.2215 & 0.2314 & 0.2289 & 0.2349 \\
NMF & 0.2556 & 0.2494 & 0.2368 & 0.2431 \\
\rowcolor{mygray}
SVD++ & 0.2493 & 0.2478 & 0.2381 & 0.2439\\
HSR & 0.2544 & 0.2495 & 0.2384 & 0.2448\\
\rowcolor{mygray}
HPF & 0.2584 & 0.2513 & 0.2405 & 0.2474\\
FM & 0.2527 & 0.2453 & 0.2284 & 0.2333\\
\rowcolor{mygray}
LambdaMART & 0.2372 & 0.2337 & 0.2272 & 0.2218\\
HBayes & \textbf{0.2685} & \textbf{0.2614} & \textbf{0.2478} & \textbf{0.2541}\\
\bottomrule
\end{tabular}
\end{center}
\end{table}


\section{Conclusion}
\label{sec:conclusion}
In this paper, we propose a novel generalized learning framework that learns both the entity hierarchical structure and its latent factors for building a personalized Bayesian recommender system, \emph{HBayes}. By utilizing variational Bayesian inference approach, HBayes is able to converge efficiently with few iterations.  In empirical studies, we walk through a practical case study of e-commerce apparel data and another publicly available music recommendation data.  Experiment results show that HBayes beats other \emph{state-of-the-art} recommendation approaches in the sense of precisions, recalls, as well as NDCG metrics, due to the fact that HBayes is able to extract the data characteristics as well as capture user hidden preferences.   

\bibliographystyle{IEEEtran}
\bibliography{bigdata2019}

\end{document}